\documentclass[letterpaper,journal]{IEEEtran}

\usepackage{graphicx}
\usepackage{amsmath}
\usepackage{amsfonts}
\usepackage{amssymb}
\usepackage{booktabs}
\usepackage{array}
\usepackage{xcolor}
\usepackage{tikz}
\usepackage{algorithm}
\usepackage{algpseudocode}
\usepackage{url}
\usepackage{textcomp}
\usepackage{stfloats}
\usepackage{cite}

\begin{document}

\title{ADM-Planner: LLM-Guided Long-Horizon Planning for Mobile Manipulators with
Attention-Enhanced Dynamic Memory
}

\author{Jiaping Xiao$^{\dagger}$,~Pingyuan~Ji$^{\dagger}$,~and~Mir~Feroskhan
\thanks{$^{\dagger}$These authors are co-first authors.}
\thanks{Jiaping Xiao, Pingyuan Ji, and Mir Feroskhan are with the the School of Mechanical and Aerospace Engineering, Nanyang Technological University, Singapore. (e-mail: jiaping001@e.ntu.edu.sg; jipi0001@e.ntu.edu.sg; mir.feroskhan@ntu.edu.sg). \textit{(Corresponding author: Mir Feroskhan.)}}}

\maketitle

\begin{abstract}

Large language models can decompose mobile-manipulation goals into long action
sequences, but the resulting plans remain reliable only while their world
context is current. A fixed scene description becomes stale when objects are
discovered, moved, or completed while retaining every observation instead produces a growing history with redundant and conflicting state. To resolve this tension, we present an LLM-guided planning framework ADM-Planner with attention-enhanced dynamic memory (ADM). Persistent workspace knowledge is separated from object-centric state, asynchronous observations and action outcomes update that state, and a bounded retriever exposes only the entries that can affect the next decision. The LLM replans when an update invalidates the remaining plan. Across
1,500 task-simulator episodes, the proposed ADM achieved 100\% full-task
success in the 14-container noisy dynamic setting, compared with 62\% for
static memory and 97\% for unfiltered dynamic memory, while reducing the
context-size proxy by 95.8\% relative to the latter. In a six-episode live
GPT-5 Mini planner, both dynamic memory variants completed every mission, while ADM reduced provider-reported input tokens by 14.4\% and mean planner calls from 7.0 to 6.0. A separate 60-trial PyBullet study retained 100\% success for ADM, compared with 50\% for static memory. Finally, the mobile manipulator with ADM-Planner completed various missions in indoor and outdoor physical experiments while incorporating targets revealed after execution began. The results show that selective state maintenance with ADM, rather than prompt history alone, is a practical basis for long-horizon planning in changing environments.
\textit{Project page: \url{https://xjp99v5.github.io/ADM-Planner/}}

\end{abstract}


\begin{figure*}[t]
    \centering
    \includegraphics[width=0.95\textwidth]{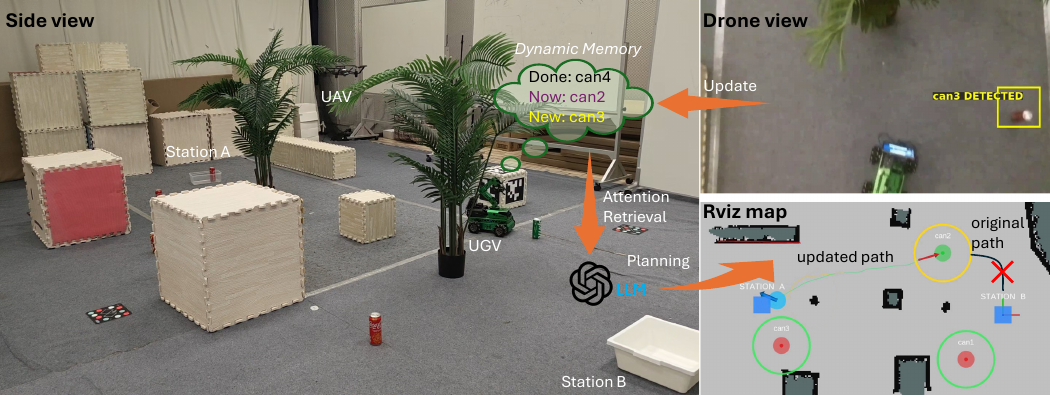}
    \caption{Dynamic aerial-to-ground memory update during a physical recycling mission. Left: the mobile manipulator operates among containers, collection stations, and obstacles in a previously unseen
    workspace. Upper right: the aerial robot detects the task-relevant
    \texttt{can3} during execution. The accepted observation updates
    mobile manipulator's memory status and LLM subsequently conducts replanning to deliver \texttt{can2} to \texttt{Station A} accordingly. Lower right: the map incorporates the updated planning information and generates a new trajectory.}
    \label{fig:cover}
\end{figure*}

\section{Introduction}

\IEEEPARstart{L}{ong-horizon} mobile manipulation couples navigation, perception, and physical
interaction over a sequence of state-dependent decisions. In container
recycling, for example, each delivery changes the set of active objects, and
objects outside the robot's local field of view may be discovered or moved while
the mission is running. A planner must therefore preserve progress over many
actions without assuming that the initial scene remains correct
\cite{shridhar2020alfred,yenamandra2023homerobot}.

Large language models (LLMs) provide a flexible mechanism for translating task
instructions into robot skills \cite{huang2022zeroshot,ichter2023saycan,
liang2023code,singh2023progprompt}. Existing systems ground these decisions in
affordances, execution feedback, maps, or scene graphs
\cite{huang2023innermonologue,rana2023sayplan}. However, their plan quality is
bounded by the state supplied at each query. Closed-loop execution does not by
itself solve the context problem when the planner can react only to information that is represented accurately and retrieved at the right time.

This creates a tension between freshness and selectivity. Keeping only the
initial map gives a compact prompt but leaves the planner acting on stale object
poses and task status. Appending every observation preserves changes but causes
the context to grow with mission duration and retains mutually inconsistent
records. The underlying challenge is thus not simply to give the LLM more
history, but to maintain a compact decision state that changes with the world.

To address this challenge, we introduce an attention-enhanced dynamic-memory planning (ADM-Planner) framework for long-horizon mobile manipulation. Stable workspace knowledge and
robot skills are stored separately from object-centric dynamic state.
Asynchronous observations and action outcomes revise the current entity records,
while a non-parametric attention score combines source-weighted confidence,
recency decay, and distance to the active task. Only the top-$K$ current records
are exposed to the LLM. A monitored executor then
continues the current plan when it remains valid and replans from the achieved
state when an update changes an unexecuted action. Figure~\ref{fig:cover}
illustrates this loop in the air-ground can recycling system.

The main contributions are the following. First, we formulate long-horizon LLM
planning as joint state maintenance and bounded retrieval, implemented through
separate static and object-centric dynamic memory. Second, we introduce an
update-retrieve-replan algorithm named ADM-Planner that reacts to asynchronous
evidence without replaying the complete interaction history. Third, we realize
the framework in an aerial-ground recycling system, where air detections
extend the mobile manipulator's task state during execution. We demonstrate the capability of the system in real-world indoor and outdoor recycling scenarios.

\section{Related Work}

LLM-based robot planners translate language goals into action sequences, skill
choices, or executable programs \cite{huang2022zeroshot,ichter2023saycan,
liang2023code,singh2023progprompt}. Feedback-driven methods improve grounding by
feeding observations and action outcomes back to the planner
\cite{huang2023innermonologue}, while scene-graph and integrated
mobile-manipulation systems connect language reasoning to larger workspaces
\cite{rana2023sayplan,liu2024okrobot,qiu2025wildlma}. These methods establish
the value of closed-loop planning, but they leave open which accumulated state
should be retained and exposed during a changing long-horizon mission.

Semantic maps and scene graphs provide persistent memory for multi-object
navigation and manipulation \cite{wani2020multion,chaplot2020objectgoal,
jatavallabhula2023conceptfusion,werby2024hovsg}. Long-term embodied agents also
retrieve prior observations to support extended reasoning
\cite{wang2026longterm}. A fixed representation is effective for stable
geometry, whereas dynamic tasks additionally require object locations,
confidence, and completion status to be revised. Our method focuses on this
interface between memory maintenance and planning, where each entity has one current task record, and retrieval is bounded by the decision rather than by mission length.

Heterogeneous aerial-ground systems use elevated sensing to share maps or support
ground navigation under limited visibility \cite{li2024colag,
cladera2025airground}. We use aerial perception as an asynchronous source of
task evidence rather than as a separate task planner. This distinction makes the
memory mechanism independent of a particular sensor platform, hence aerial detections, onboard observations, and skill outcomes all enter the same update and
plan-validity process.

\section{Problem Formulation}

Let $g$ denote a long-horizon manipulation goal and $x_t$ the robot execution
state at time $t$. The planner has access to static memory $M^s$, containing the
workspace map, station poses, task rules, and available skills, and dynamic
memory $M_t^d$, containing the current state of task-relevant entities:
\begin{equation}
M_t^d=\{e_i^t\}_{i=1}^{N_t}, \quad
e_i^t=(z_i,c_i,p_i,q_i,\sigma_i,\tau_i,o_i),
\label{eq:memory-entry}
\end{equation}
where $z_i$ is an entity identifier, $c_i$ its class, $p_i$ its estimated pose,
$q_i$ its task status, $\sigma_i$ its confidence, $\tau_i$ its latest update
time, and $o_i$ the observation source. Status progresses through
\textit{observed}, \textit{assigned}, \textit{grasped},
\textit{delivered}, or \textit{failed}.

At each planning event, the system must choose a bounded subset
$R_t\subseteq M_t^d$, $|R_t|\leq K$, and generate a feasible skill sequence
\begin{equation}
P_t=f_\theta(g,x_t,M^s,R_t),
\label{eq:planner}
\end{equation}
where $f_\theta$ is the LLM planner. The subset must include state that can
change the next decision while excluding stale, completed, and irrelevant
records. The objective is to complete the mission while limiting failed actions,
unnecessary travel, replanning calls, and context cost. Low-level collision
avoidance, trajectory generation, and grasp control remain outside the LLM but are implemented by the reliable low-level controller. The
planner only invokes skills exposed by the robot interface.

\section{Methodology}
\label{sec:method}

\subsection{Overview}

The ADM-Planner framework closes the loop between state estimation and high-level planning.
Each iteration comprises three stages. First, an event interface converts
asynchronous perceptual reports and synchronous skill outcomes into updates of a
single object-centric world state. Second, an attention-based retriever filters terminal or
unsupported records and serializes at most $K$ scored entities. Third, the LLM
proposes a typed skill sequence that is validated before one skill is dispatched.
Outcomes return through the event interface, so later plans retain achieved
progress. Perception may update memory whenever it arrives, but planning occurs
only when no plan exists, execution fails, or an update affects the remaining
plan. Thus, the LLM selects task-level actions without directly modifying memory
or commanding actuators. Algorithm~\ref{alg:memory-planning} summarizes the loop.

\subsection{Object-Centric State Updates}

The split between $M^s$ and $M_t^d$ prevents persistent scene knowledge from
being rewritten whenever an object changes. Static memory stores the map,
collection-station poses, semantic regions, task rules, and skill definitions.
Dynamic memory stores one current record per task entity. A grasp changes an
object from \textit{assigned} to \textit{grasped}; placement changes it to
\textit{delivered}; and a new observation may revise its pose and confidence.
Earlier values can remain in an audit log, but only the current record is
eligible for planning.

Every observation is transformed into the shared map frame and normalized to
$d_k=(\hat c_k,\hat p_k,\hat\sigma_k,\hat\tau_k,\hat o_k)$, where the hatted
variables denote the measured class, pose, confidence, timestamp, and source.
For the set of class-compatible records
$\mathcal A_k=\{i:c_i=\hat c_k\}$, association selects
\begin{equation}
\begin{aligned}
j^*&=\arg\min_{i\in\mathcal A_k}\lVert p_i-\hat p_k\rVert_2,\\
m_k&=\mathbf 1[\lVert p_{j^*}-\hat p_k\rVert_2\leq r_{\mathrm{assoc}}].
\end{aligned}
\label{eq:data-association}
\end{equation}
Here $j^*$ is the nearest class-compatible record, $m_k$ is the binary match
indicator, $\mathbf 1[\cdot]$ is the indicator function, and
$r_{\mathrm{assoc}}$ is the association radius. If $\mathcal A_k$ is empty or
$m_k=0$, a new identifier is allocated; otherwise the matched record is updated.
The gate preserves class identity while allowing unseen targets to enter memory
during execution.

Update arbitration is field-specific. Accepted perceptual evidence will revise
pose, confidence, time, and source, while task status is advanced by verified
skill outcomes. Confidence and recency resolve competing estimates, but an
observation cannot demote a verified grasp or delivery. A failed skill marks the
entity for reconsideration without discarding completed work. Aerial and onboard
detections and skill outcomes all use this update interface.

After confidence and pose-tolerance filtering, each update yields changed fields
$\mathcal C_t=\{(z_i,h):e_{i,h}^{t}\neq e_{i,h}^{t-1}\}$, where
$h\in\{c,p,q,\sigma,\tau,o\}$ denotes a mutable record field and $e_{i,h}^{t}$
is the value of field $h$ in entity $z_i$ at time $t$. Let $D(P_t)$ contain the
fields used by arguments and preconditions in the remaining plan. Replanning is
required exactly when $\mathcal C_t\cap D(P_t)\neq\emptyset$. A revised target
pose therefore invalidates navigation to it, whereas an unrelated detection does
not interrupt the active skill. This test decouples observation and LLM-query
rates.

\subsection{Attention-Enhanced Dynamic Memory Retrieval}

The ADM maintains a separate score over candidate locations for each active
entity. Let $j$ be the record matched to observation $d_k$ by
Eq.~\eqref{eq:data-association} (or a newly allocated record). For an accepted
positive observation at time $t$, the update is
\begin{equation}
\begin{aligned}
\tilde b_j^t(p)&=\exp\!\left(-\frac{t-\tau_j}{\beta}\right)
b_j^{\tau_j}(p),\\
b_j^t(p)&=\tilde b_j^t(p)+w(\hat o_k)\hat\sigma_k
\mathbf 1[p=\hat p_k],
\end{aligned}
\label{eq:belief-update}
\end{equation}
where $b_j^t(p)$ is the support for entity $j$ being at candidate pose $p$,
$\beta$ is the decay horizon, and $w(\hat o_k)$ encodes source reliability.
Thus, every previous pose hypothesis is first discounted, while new evidence is
added only at $\hat p_k$; repeated consistent observations accumulate, whereas
an unsupported old pose eventually vanishes. Remote observations below the
confidence threshold are rejected before this update. A verified onboard
observation instead resets the belief to the measured pose, and a verified
absence deletes that candidate. Scores below a small support threshold are
pruned.

At a planning event, delivered and carried entities are first excluded. For
each remaining entity, Eq.~\eqref{eq:belief-update}'s decayed belief is reduced
to one planner record using $\hat p_i=\arg\max_p\tilde b_i^t(p)$ and
$\bar\sigma_i(t)=\min\{1,\max_p\tilde b_i^t(p)\}$. The former supplies the target
pose and the latter supplies the confidence in the attention score
\begin{equation}
s_i(t)=\frac{\bar\sigma_i(t)}
{1+\lVert p_t^r-\hat p_i\rVert_1},
\label{eq:attention}
\end{equation}
where $p_t^r$ is the robot position. This score favors reliable nearby targets
while recency is already encoded in $\bar\sigma_i(t)$ by
Eq.~\eqref{eq:belief-update}. The planning context is the bounded set
\begin{equation}
R_t=\operatorname{Top}_K\{e_i^t:i\text{ active},\;
\bar\sigma_i(t)\geq\sigma_{\min}\},
\label{eq:retrieval}
\end{equation}
ordered by $s_i(t)$. This attention-enhanced mechanism is to restrict the planner input to current,
decision-relevant object estimates. Because the prompt contains only the
selected current record for each entity, its dynamic portion is bounded by
$K$ even as observations accumulate. Static rules and completion summaries are
serialized separately and do not compete for retrieval slots.

\begin{algorithm}[t]
\caption{Attention-enhanced dynamic-memory planning}
\label{alg:memory-planning}
\begin{algorithmic}[1]
\Require Goal $g$, static memory $M^s$, dynamic memory $M_0^d$, capacity $K$
\State $P\gets\emptyset$; observe initial robot state $x$
\While{the completion condition for $g$ is false}
    \State Receive observations $\mathcal O_t$ and skill outcome $y_t$
    \State $(M_t^d,\mathcal C_t)\gets\Call{Update}{M_{t-1}^d,\mathcal O_t,y_t}$ with Eq.~\eqref{eq:belief-update}
    \If{$P=\emptyset$ \textbf{or} $\Call{Affects}{\mathcal C_t,P}$ \textbf{or} $\Call{Failed}{y_t}$}
        \State Form query $u_t$ from $g$, $x$, and achieved progress
        \State Score active entries with Eq.~\eqref{eq:attention}
        \State $R_t\gets\Call{BoundedRetrieve}{M_t^d,u_t,K}$
        \State $P\gets\Call{Validate}{f_\theta(g,x,M^s,R_t)}$
    \EndIf
    \State $a\gets\Call{PopFirst}{P}$
    \State $(x,y_{t+1})\gets\Call{Execute}{a}$
\EndWhile
\Ensure All required entities satisfy their terminal task status
\end{algorithmic}
\end{algorithm}

\subsection{LLM-Guided Skill Planning and Execution}

At a planning event, the system serializes a query $u_t$ with five blocks: (i)
the mission goal and completion condition; (ii) the robot pose, current skill
outcome, and held object; (iii) static task rules, station assignments, and the
available skill signatures; (iv) the retrieved active records $R_t$; and (v) a
compact summary of delivered or otherwise terminal entities. This separation
lets the LLM reason over the current decision state without receiving the full
observation history. Entity identifiers and poses are copied from memory rather
than inferred from free-form scene text.

Conditioned on $u_t$, the LLM performs high-level task decomposition. It selects
an eligible target, identifies its compatible station, and orders the skills
needed to complete that subgoal while respecting the current execution state.
For example, an unheld observed container may require
\texttt{navigate}$\rightarrow$\texttt{approach}$\rightarrow$\texttt{grasp},
followed by navigation and \texttt{deliver}; if an object is already held, the
pickup prefix is omitted. The output is the finite sequence
$P_t=(a_1,\ldots,a_H)$, where $a_j=(\ell_j,\vartheta_j)$ contains a skill name
and typed arguments. The output space is restricted to exposed skills such as
\texttt{navigate}, \texttt{approach}, \texttt{grasp}, \texttt{deliver}, and
\texttt{verify}; the LLM neither generates actuator commands nor directly edits
memory.

The generated sequence is grounded before execution. A deterministic validator
checks skill names, argument presence and types, entity references, carrying
capacity, target status, and object--station compatibility. Invalid output is
rejected rather than dispatched. For a valid plan, the executor rechecks the
next action against the latest memory and invokes the corresponding
low-level controller, which handles trajectory
generation, navigation, collision avoidance, and grasp control.

Execution is receding-horizon at the skill level: only the next validated skill
is committed, and its observed outcome is written to $M_t^d$ before the
following skill. The remaining suffix is retained while its preconditions hold,
but is cleared when execution fails or
$\mathcal C_t\cap D(P_t)\neq\emptyset$. The next LLM query then starts from the
achieved state. Thus, delivered containers remain complete, newly observed
targets become candidates, and replanning repairs the updated future part
of the mission.

\begin{figure*}[!t]
    \centering
    \includegraphics[width=0.98\textwidth]{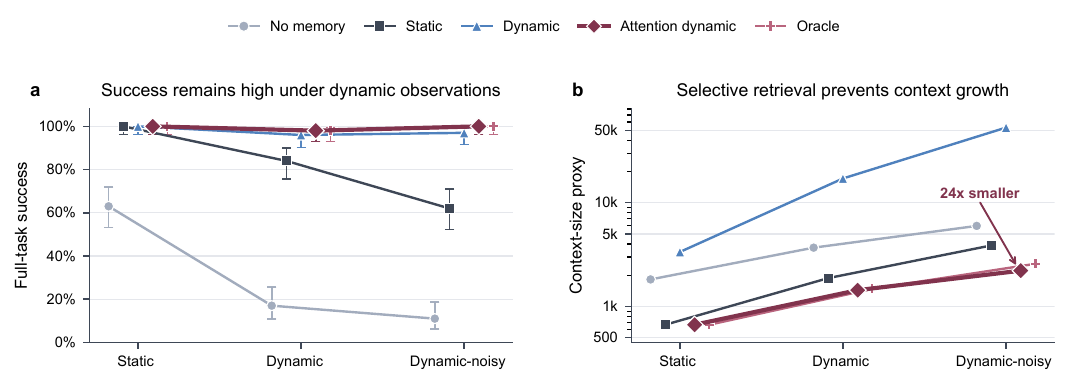}
    \caption{Task-level memory comparison over 100 paired seeds per scenario.
    (a) Full-task success with Wilson 95\% confidence intervals. (b)
    Accumulated context-size proxy on a logarithmic scale. Attention-enhanced
    memory approaches oracle success while avoiding unfiltered history growth.}
    \label{fig:task-simulation}
\end{figure*}

\section{Experiments}
\label{sec:experiments}

The experiments proceed from controlled simulation to physical full-system validation. We first isolate the
core mechanism, test it with the deployed planner, increase execution fidelity,
and finally validate the complete physical system. We ask whether dynamic updates
recover from stale state, whether bounded retrieval preserves that benefit
without unbounded context, and whether the resulting plans remain executable.

\begin{table}[!t]
\caption{Task-level success over 100 paired seeds per scenario. Context is the mean proxy for
Dynamic-noisy-14.}
\label{tab:task-simulation}
\centering
\setlength{\tabcolsep}{3.5pt}
\begin{tabular}{lrrrr}
\toprule
Method & S-6 $\uparrow$ & D-10 $\uparrow$ & DN-14 $\uparrow$ & Context $\downarrow$ \\
\midrule
No memory~\cite{huang2022zeroshot} & 63 & 17 & 11 & 5{,}978 \\
Static memory~\cite{chaplot2020objectgoal} & 100 & 84 & 62 & 3{,}876 \\
Unfiltered dynamic~\cite{wang2026longterm} & 100 & 96 & 97 & 53{,}049 \\
\textbf{ADM (ours)} & \textbf{100} & \textbf{98} & \textbf{100} & \textbf{2{,}212} \\
Oracle state & 100 & 98 & 100 & 2{,}576 \\
\bottomrule
\end{tabular}
\end{table}

\begin{figure*}[!t]
    \centering
    \resizebox{0.98\textwidth}{!}{\input{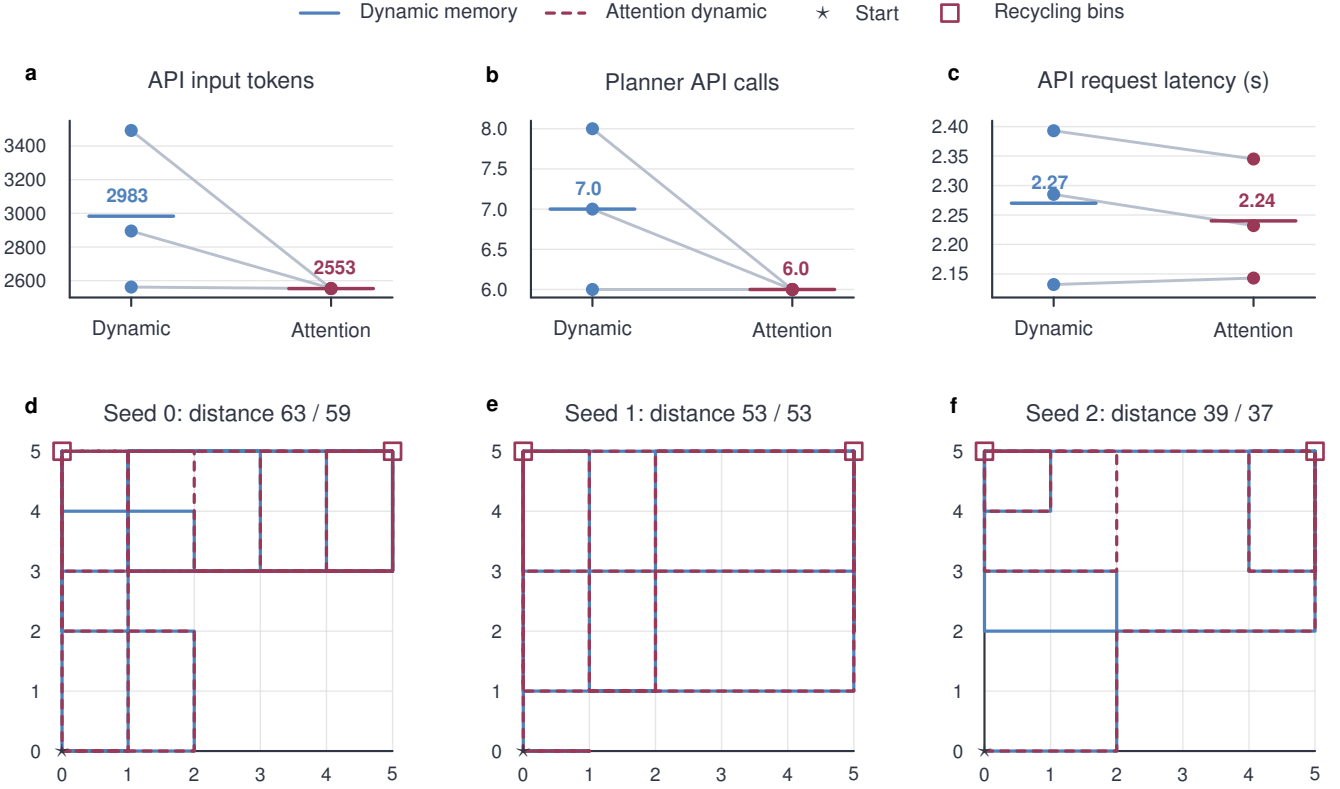}}
    \caption{Live GPT-5 Mini pilot and exact trajectory replays for three paired
    seeds per method. (a)--(c) Gray segments join matched seeds; horizontal
    ticks show means for input tokens, API calls, and latency. (d)--(f) Replayed
    $6\!\times\!6$ robot paths. Titles report Dynamic/Attention travel
    distance; the star is the start, squares are stations, and crosses are
    terminal positions.}
    \label{fig:llm-pilot}
\end{figure*}

\subsection{Task-Level Memory Evaluation}

\textbf{Setup.} We use the grid-based semantic simulator to isolate memory and
replanning from control. It provides unit-cost \texttt{MOVE}, \texttt{PICK},
and \texttt{PLACE} actions, one-object capacity, type-specific stations,
delayed reports, relocation, and injected failures. The three scenarios use
6 (S-6), 10 (D-10), and 14 (DN-14) containers on $5\!\times\!5$, $7\!\times\!7$, and
$8\!\times\!8$ grids with 80, 150, and 240-step budgets. Across scenarios,
relocation increases from 0 to 0.025, false reports from 0.05 to 0.30, and
latency from two to four steps; detection decreases from 0.85 to 0.68.

We compare no memory, initial static memory, unfiltered dynamic memory, the
proposed attention-enhanced dynamic memory, and oracle state. Attention uses
$K=8$, a 0.55 remote-observation acceptance threshold, and an 18-step recency
decay. Each method receives identical event tapes for 100 seeds per scenario
(1,500 episodes total). A deterministic nearest confidence-weighted policy
isolates memory effects. Context cost is $48+18n_t$ per planning call, where
$n_t$ is the number of exposed records; this is a size proxy, not tokenization.
We use Wilson success intervals and bootstrap paired-difference intervals.

\begin{figure*}[!t]
    \centering
    \includegraphics[width=0.94\textwidth]{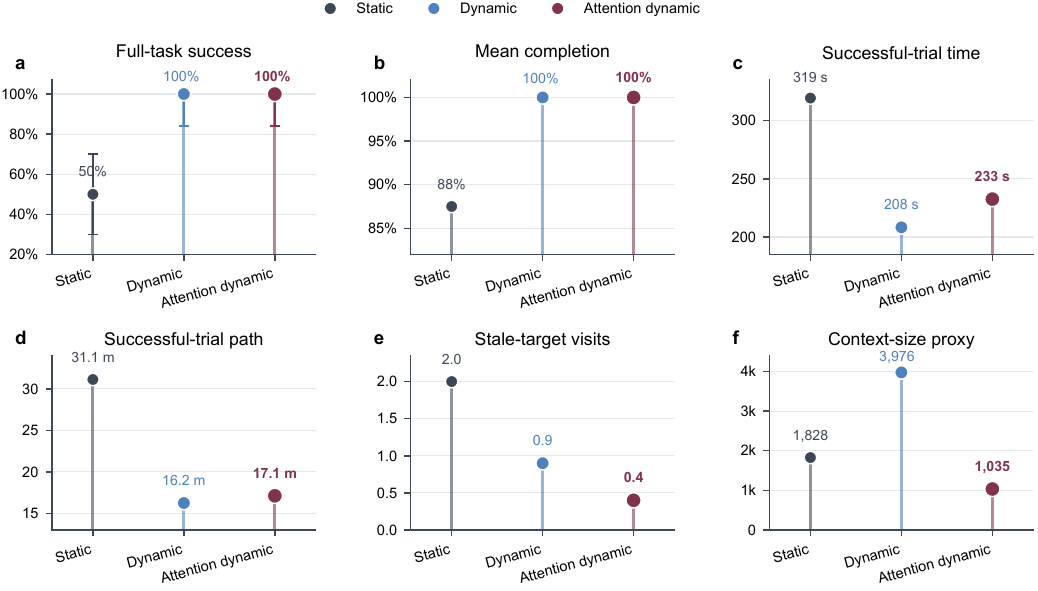}
    \includegraphics[width=0.94\textwidth]{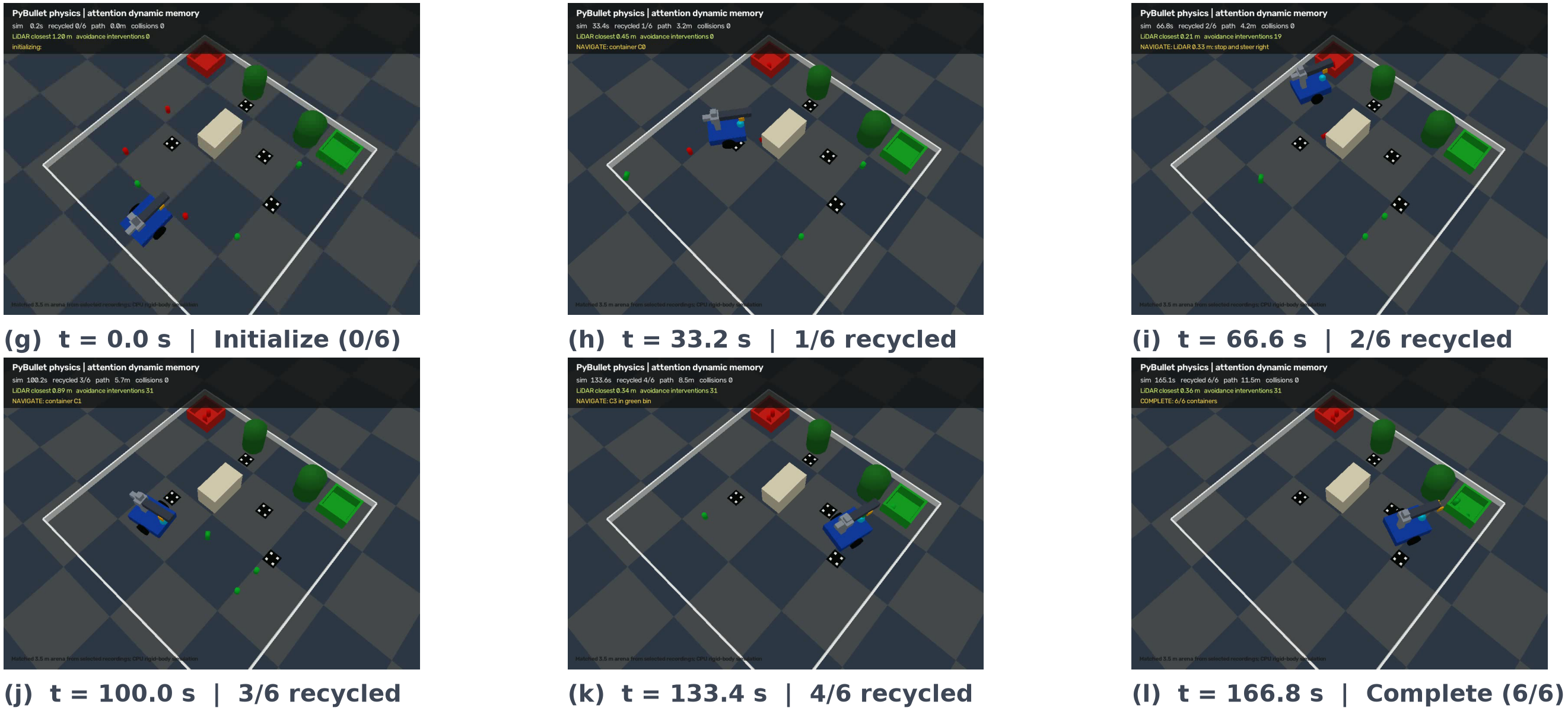}
    \caption{Rigid-body validation over 20 paired seeds per method: (a)
    full-task success, (b) completion, (c) successful-trial time, (d)
    successful-trial path length, (e) stale-target visits, and (f) context-size
    proxy. Error bars in (a) are Wilson 95\% confidence intervals. (g)--(l)
    Time-sequential views of a representative attention-memory rollout (seed
    6), progressing from initialization to six completed deliveries over
    166.8~s of recorded video.}
    \label{fig:physics-simulation}
\end{figure*}

\textbf{Results.} Table~\ref{tab:task-simulation} and
Fig.~\ref{fig:task-simulation} show that in DN-14, static,
unfiltered dynamic, and attention memory achieve 62\%, 97\%, and 100\% success,
respectively. Attention gains 38 points over static memory (95\% CI: 28--48),
while its three-point gain over unfiltered memory is inconclusive (95\% CI:
0--7). It also reduces context from 53,049 to 2,212 units (95.8\%) and episode
length from 209.7 to 201.3 steps (paired saving 8.4; 95\% CI: 6.4--10.7), while
matching oracle success. The static-to-dynamic gain shows that state revision is
necessary when reports arrive late or objects move, whereas the context reduction
shows that retaining every revision is unnecessary for the next decision.
However, Eq.~\eqref{eq:attention} is a hand-designed confidence--distance
heuristic and may suppress a distant entity important to a future subgoal. Fixed
$K$, source weights, and confidence and decay thresholds require calibration;
their individual effects remain to be ablated.

\subsection{Live LLM Planning and Trajectory Analysis}

\textbf{Setup.} We next replace the policy surrogate with the frozen
\texttt{gpt-5-mini-2025-08-07} snapshot through the OpenAI Responses API. Each
request uses strict structured output, no provider-side storage, and only
retrieved records. The $6\!\times\!6$, six-container scenario has a 110-step
budget, 0.02 relocation, 0.80 detection, 0.25 false-report probability, and
two-step latency. Attention uses $K=6$, a 0.55 threshold, and 16-step decay. We
run three paired seeds per dynamic variant with identical event tapes.
An abbreviated exchange from the first attention-memory episode was:
\begin{quote}
\footnotesize\ttfamily
Input: task=select\_next\_container; robot=[0,0];\newline
retrieved=[\{id:0, red, pose:[2,2], bin:[0,5], conf:1.0\},
\{id:5, green, pose:[1,3], bin:[5,5], conf:1.0\}, $\ldots$];\newline
skills=[navigate\_to\_container, align\_base, extend\_arm, grasp,
retract\_arm, navigate\_to\_bin, place\_inside\_bin].\newline
Output: \{id:0, pose:[2,2], bin:[0,5], actions:[navigate\_to\_container,
align\_base, extend\_arm, grasp, retract\_arm, navigate\_to\_bin,
align\_base, extend\_arm, place\_inside\_bin, retract\_arm],
reason: closest high-confidence target\}.
\end{quote}

\textbf{Results.} Across the six live GPT-5 Mini episodes, both dynamic variants
complete every mission. ADM reduces mean provider-reported input tokens from
2,983 to 2,553 (14.4\%), total tokens from 3,853 to 3,294 (14.5\%), and planner
calls from 7.0 to 6.0. Mean request latency changes only slightly, from 2.27 to
2.24~s. We therefore
treat latency as comparable rather than claim a runtime benefit.
Figure~\ref{fig:llm-pilot} replays the recorded decisions exactly. Because both
variants use the same structured schema, the token reduction confirms that
bounded retrieval remains effective under provider tokenization rather than only
under the context-size proxy.

\subsection{Rigid-Body Execution Validation}

\textbf{Setup.} We test whether the task-level plans remain executable under
rigid-body dynamics using PyBullet 3.2.7 in headless DIRECT mode at 120~Hz. The
3.5-m arena contains a differential-drive base, articulated arm, collisions,
gravity, contact-gated grasps, two type-specific stations, six containers, and
scheduled relocations. We run 20 paired seeds for static, unfiltered dynamic,
and ADM.

\begin{table}[t]
\caption{PyBullet validation over 20 paired seeds per method.}
\label{tab:physics-simulation}
\centering
\setlength{\tabcolsep}{4.0pt}
\begin{tabular}{lrrr}
\toprule
Method & Success (\%) $\uparrow$ & Completion (\%) $\uparrow$ & Context $\downarrow$ \\
\midrule
Static memory & 50 & 87.5 & 1{,}828 \\
Unfiltered dynamic & \textbf{100} & \textbf{100.0} & 3{,}976 \\
\textbf{ADM (ours)} & \textbf{100} & \textbf{100.0} & \textbf{1{,}035} \\
\bottomrule
\end{tabular}
\end{table}

\begin{figure*}[!t]
    \centering
    \includegraphics[width=0.96\textwidth]{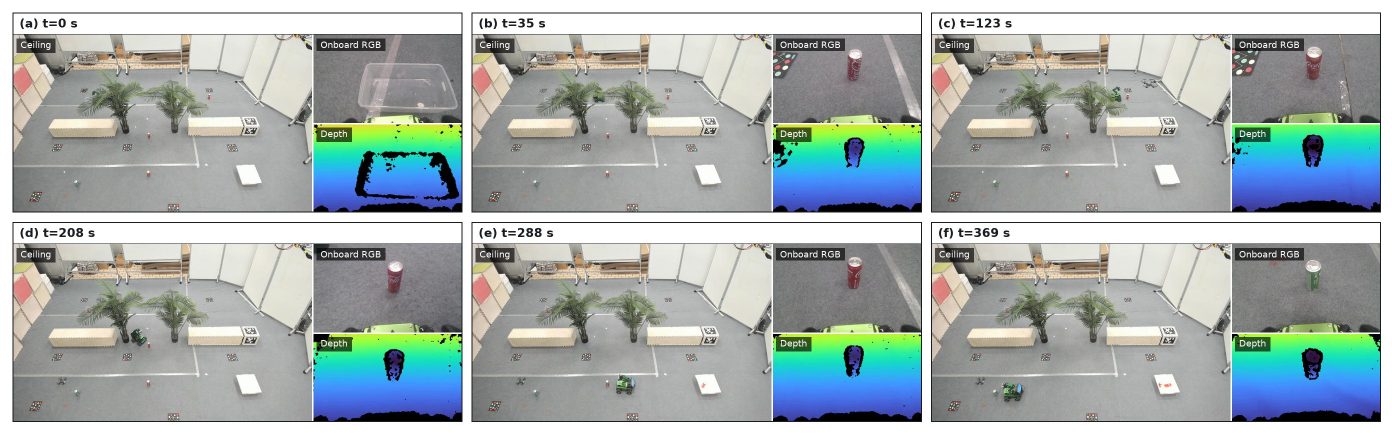}\\[0.5mm]
    \includegraphics[width=0.96\textwidth]{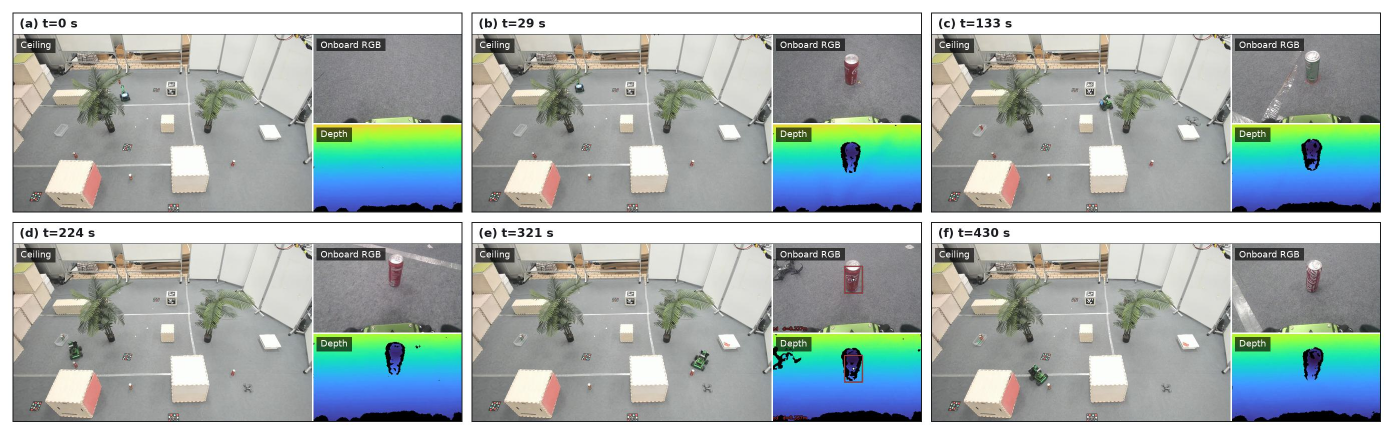}\\[0.5mm]
    \includegraphics[width=0.96\textwidth]{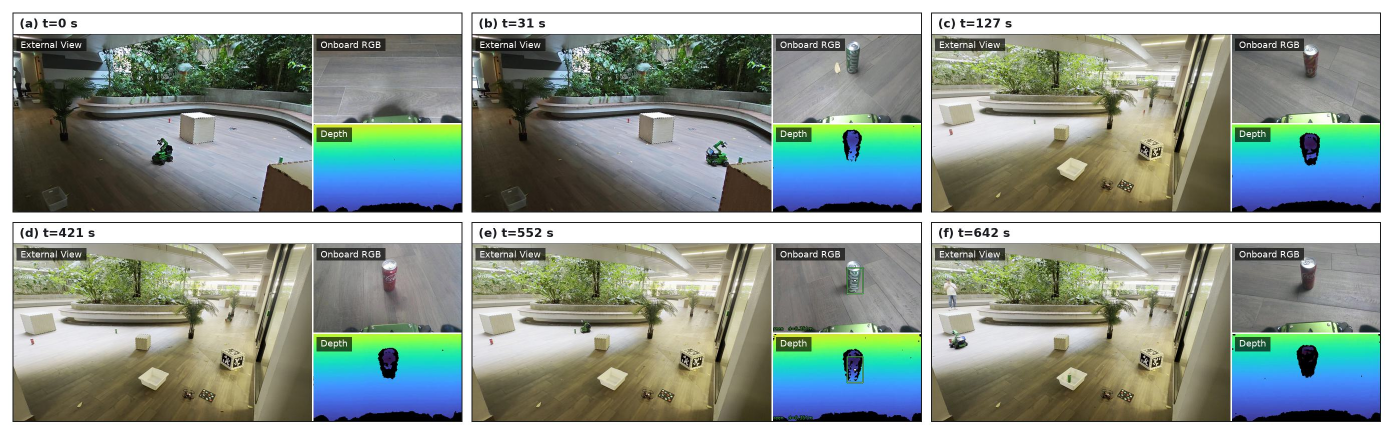}\\[0.5mm]
    \includegraphics[width=0.96\textwidth]{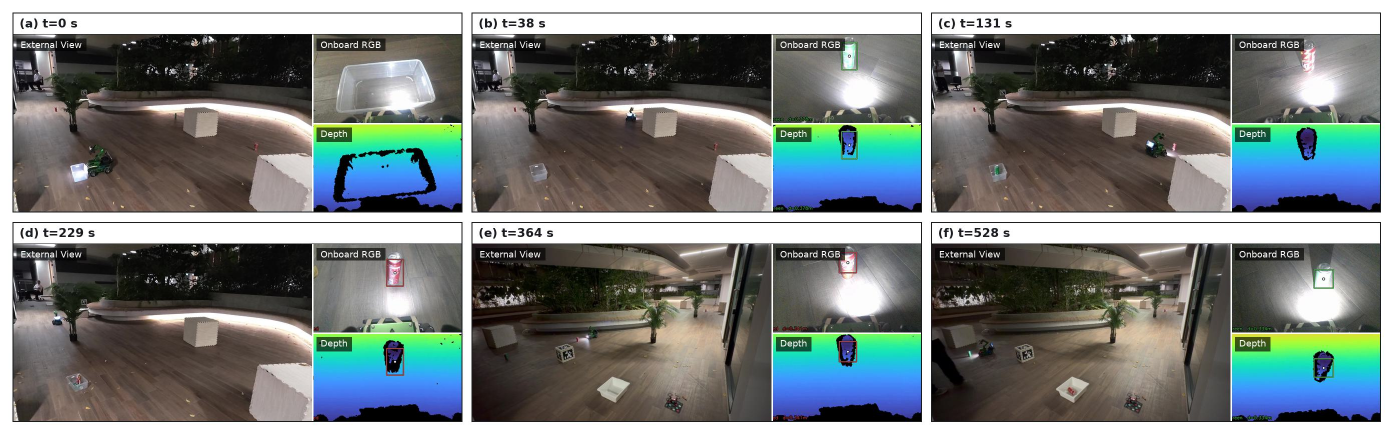}\\[0.5mm]
    \caption{Physical recycling experiments. From top to bottom, the rows show
    the narrow-gate scenario, cluttered scenario, unstructured daytime scenario,
    and nighttime scenario. Every run delivered five containers to two stations with dynamic memory updates.}
    \label{fig:physical-configurations}
\end{figure*}

\textbf{Results.} Both dynamic variants complete all 20 trials (100\%; 95\%
CI: 83.9--100\%), compared with 10 of 20 for static memory (50\%;
29.9--70.1\%). Attention reduces stale-target visits from 0.9 to 0.4 and the
context proxy from 3,976 to 1,035 (74.0\%). Its successful-trial time/path are
232.6~s/17.1~m versus 208.3~s/16.2~m for unfiltered memory, so attention does
not improve every execution cost. Thus, bounded retrieval targets informational
efficiency and stale-state failures rather than motion optimality. The preserved
success ordering shows that task-level plans remain executable with contacts,
gravity, and collision constraints, but the approximate arena supports
feasibility rather than calibrated sim-to-real prediction.

\subsection{Physical System Validation}

\textbf{Setup.} The physical system consists of a JetRover mobile
manipulator~\cite{hiwonder2026jetrover}, an Tello drone, beverage containers,
obstacles, and two stations. An onboard Jetson Orin computer runs ROS Noetic
perception and control; a RealSense D435i RGB-D camera supports grasping, and 2-D
LiDAR with HECTOR SLAM~\cite{kohlbrecher2011hector} provides mapping and
localization. The aerial robot reports detections in the shared map frame.
When an accepted memory update affects the remaining plan, the system replans
using the same frozen GPT-5 Mini and OpenAI Responses API.
The omnidirectional base uses HECTOR scan matching and a TEB local planner to
navigate to a collision-checked standoff pose. It then holds yaw fixed and uses
RGB-D feedback to translate longitudinally and laterally until the target is
centered about 0.35~m ahead. Color segmentation and depth back-projection
estimate the target in the arm frame; inverse kinematics executes a pre-grasp,
descent, gripper closure, and lift, and the gripper-servo position verifies the
capture before transport to the assigned station.

Each mission contains five containers where one is initially known and four are
revealed online. Configurations vary object poses, obstacles, connectivity, and
lighting conditions, and environment difficulty. The assigned station is dynamically determined for each mission based on the initial configuration and online reveals. 
Completion requires all containers being placed at their assigned stations and no held object.

\textbf{Results.} All configurations complete the container collection task while incorporating online reveals. Thus, the 
update-replan loop functions across the layouts in
Fig.~\ref{fig:physical-configurations}, exercising the complete path from aerial
detection through memory revision to ground execution rather than replaying a
fully known scene. More details are provided in the supplementary video. These demonstrations establish feasibility, and adaptiveness to
various layouts and lighting conditions, detector uncertainty, or communication interruptions. Localization error may create duplicate or misplaced records. 
Future studies should evaluate uncertainty-aware association, active observation requests, communication loss, and longer missions.

\section{Conclusion}

This paper addressed the memory problem in long-horizon LLM planning, where a robot
needs state that is current enough to react to change and selective enough to
remain useful for the next decision. The proposed framework separates persistent
knowledge from object-centric dynamic state, retrieves a bounded set of
plan-relevant records, and replans only when new evidence affects the remaining
skill sequence. The controlled simulation shows that dynamic updates recover the
failures of static memory, while attention removes most of the context growth of
unfiltered history. Live-LLM, rigid-body, and physical experiments then verify
the same mechanism at increasing levels of realism. Thus, these results
support attention-enhanced dynamic memory as a practical interface between
asynchronous perception and long-horizon mobile-manipulator planning.


\bibliographystyle{IEEEtranBST/IEEEtran}
\bibliography{IEEEtranBST/IEEEabrv,references}

@inproceedings{chaplot2020objectgoal,
  author    = {Chaplot, Devendra Singh and Gandhi, Dhiraj Prakashchand and Gupta, Abhinav and Salakhutdinov, Ruslan},
  title     = {Object Goal Navigation Using Goal-Oriented Semantic Exploration},
  booktitle = {Advances in Neural Information Processing Systems},
  volume    = {33},
  pages     = {4247--4258},
  year      = {2020},
}

@article{cladera2025airground,
  author  = {Cladera, Fernando and Ravichandran, Zachary and Hughes, Jason and Murali, Varun and Nieto-Granda, Carlos and Hsieh, M. Ani and Pappas, George J. and Taylor, Camillo J. and Kumar, Vijay},
  title   = {Air-Ground Collaboration for Language-Specified Missions in Unknown Environments},
  journal = {{IEEE} Trans. Field Robot.},
  volume  = {2},
  pages   = {626--642},
  year    = {2025},
  doi     = {10.1109/TFR.2025.3584019},
}

@inproceedings{huang2022zeroshot,
  author    = {Huang, Wenlong and Abbeel, Pieter and Pathak, Deepak and Mordatch, Igor},
  title     = {Language Models as Zero-Shot Planners: Extracting Actionable Knowledge for Embodied Agents},
  booktitle = {Proceedings of the 39th International Conference on Machine Learning},
  series    = {Proceedings of Machine Learning Research},
  volume    = {162},
  pages     = {9118--9147},
  publisher = {PMLR},
  year      = {2022},
}

@inproceedings{huang2023innermonologue,
  author    = {Huang, Wenlong and Xia, Fei and Xiao, Ted and Chan, Harris and Liang, Jacky and Florence, Pete and Zeng, Andy and Tompson, Jonathan and Mordatch, Igor and Chebotar, Yevgen and Sermanet, Pierre and Jackson, Tomas and Brown, Noah and Luu, Linda and Levine, Sergey and Hausman, Karol and Ichter, Brian},
  title     = {Inner Monologue: Embodied Reasoning Through Planning With Language Models},
  booktitle = {Proceedings of the 6th Conference on Robot Learning},
  series    = {Proceedings of Machine Learning Research},
  volume    = {205},
  pages     = {1769--1782},
  publisher = {PMLR},
  year      = {2023},
}

@inproceedings{ichter2023saycan,
  author    = {Ichter, Brian and Brohan, Anthony and Chebotar, Yevgen and Finn, Chelsea and Hausman, Karol and Herzog, Alexander and Ho, Daniel and Ibarz, Julian and Irpan, Alex and Jang, Eric and Julian, Ryan and Kalashnikov, Dmitry and Levine, Sergey and Lu, Yao and Parada, Carolina and Rao, Kanishka and Sermanet, Pierre and Toshev, Alexander T. and Vanhoucke, Vincent and Xia, Fei and Xiao, Ted and Xu, Peng and Yan, Mengyuan and Brown, Noah and Ahn, Michael and Cortes, Omar and Sievers, Nicolas and Tan, Clayton and Xu, Sichun and Reyes, Diego and Rettinghouse, Jarek and Quiambao, Jornell and Pastor, Peter and Luu, Linda and Lee, Kuang-Huei and Kuang, Yuheng and Jesmonth, Sally and Joshi, Nikhil J. and Jeffrey, Kyle and Ruano, Rosario Jauregui and Hsu, Jasmine and Gopalakrishnan, Keerthana and David, Byron and Zeng, Andy and Fu, Chuyuan Kelly},
  title     = {Do As I Can, Not As I Say: Grounding Language in Robotic Affordances},
  booktitle = {Proceedings of the 6th Conference on Robot Learning},
  series    = {Proceedings of Machine Learning Research},
  volume    = {205},
  pages     = {287--318},
  publisher = {PMLR},
  year      = {2023},
}

@inproceedings{jatavallabhula2023conceptfusion,
  author    = {Jatavallabhula, Krishna Murthy and Kuwajerwala, Alihusein and Gu, Qiao and Omama, Mohd and Iyer, Ganesh and Saryazdi, Soroush and Chen, Tao and Maalouf, Alaa and Li, Shuang and Keetha, Nikhil Varma and Tewari, Ayush and Tenenbaum, Joshua and de Melo, Celso and Krishna, Madhava and Paull, Liam and Shkurti, Florian and Torralba, Antonio},
  title     = {ConceptFusion: Open-Set Multimodal 3D Mapping},
  booktitle = {Proceedings of Robotics: Science and Systems},
  address   = {Daegu, Republic of Korea},
  year      = {2023},
  doi       = {10.15607/RSS.2023.XIX.066},
}

@inproceedings{li2024colag,
  author    = {Li, Zhehan and Mao, Rui and Chen, Nanhe and Xu, Chao and Gao, Fei and Cao, Yanjun},
  title     = {{ColAG}: A Collaborative Air-Ground Framework for Perception-Limited {UGV}s' Navigation},
  booktitle = {2024 {IEEE} International Conference on Robotics and Automation ({ICRA})},
  pages     = {16781--16787},
  year      = {2024},
  doi       = {10.1109/ICRA57147.2024.10611264},
}

@inproceedings{liang2023code,
  author    = {Liang, Jacky and Huang, Wenlong and Xia, Fei and Xu, Peng and Hausman, Karol and Ichter, Brian and Florence, Pete and Zeng, Andy},
  title     = {Code as Policies: Language Model Programs for Embodied Control},
  booktitle = {2023 {IEEE} International Conference on Robotics and Automation ({ICRA})},
  pages     = {9493--9500},
  year      = {2023},
  doi       = {10.1109/ICRA48891.2023.10160591},
}

@inproceedings{liu2024okrobot,
  author    = {Liu, Peiqi and Orru, Yaswanth and Vakil, Jay and Paxton, Chris and Shafiullah, Nur Muhammad Mahi and Pinto, Lerrel},
  title     = {Demonstrating {OK-Robot}: What Really Matters in Integrating Open-Knowledge Models for Robotics},
  booktitle = {Proceedings of Robotics: Science and Systems},
  address   = {Delft, Netherlands},
  year      = {2024},
  doi       = {10.15607/RSS.2024.XX.091},
}

@inproceedings{qiu2025wildlma,
  author    = {Qiu, Ri-Zhao and Song, Yuchen and Peng, Xuanbin and Suryadevara, Sai Aneesh and Yang, Ge and Liu, Minghuan and Ji, Mazeyu and Jia, Chengzhe and Yang, Ruihan and Zou, Xueyan and Wang, Xiaolong},
  title     = {{WildLMa}: Long Horizon Loco-Manipulation in the Wild},
  booktitle = {2025 {IEEE} International Conference on Robotics and Automation ({ICRA})},
  pages     = {10011--10019},
  year      = {2025},
  doi       = {10.1109/ICRA55743.2025.11128535},
}

@inproceedings{rana2023sayplan,
  author    = {Rana, Krishan and Haviland, Jesse and Garg, Sourav and Abou-Chakra, Jad and Reid, Ian and S{\"u}nderhauf, Niko},
  title     = {{SayPlan}: Grounding Large Language Models Using 3D Scene Graphs for Scalable Robot Task Planning},
  booktitle = {Proceedings of the 7th Conference on Robot Learning},
  series    = {Proceedings of Machine Learning Research},
  volume    = {229},
  pages     = {23--72},
  publisher = {PMLR},
  year      = {2023},
}

@inproceedings{shridhar2020alfred,
  author    = {Shridhar, Mohit and Thomason, Jesse and Gordon, Daniel and Bisk, Yonatan and Han, Winson and Mottaghi, Roozbeh and Zettlemoyer, Luke and Fox, Dieter},
  title     = {{ALFRED}: A Benchmark for Interpreting Grounded Instructions for Everyday Tasks},
  booktitle = {Proceedings of the {IEEE/CVF} Conference on Computer Vision and Pattern Recognition},
  pages     = {10740--10749},
  year      = {2020},
}

@inproceedings{singh2023progprompt,
  author    = {Singh, Ishika and Blukis, Valts and Mousavian, Arsalan and Goyal, Ankit and Xu, Danfei and Tremblay, Jonathan and Fox, Dieter and Thomason, Jesse and Garg, Animesh},
  title     = {{ProgPrompt}: Generating Situated Robot Task Plans Using Large Language Models},
  booktitle = {2023 {IEEE} International Conference on Robotics and Automation ({ICRA})},
  pages     = {11523--11530},
  year      = {2023},
  doi       = {10.1109/ICRA48891.2023.10161317},
}

@inproceedings{wani2020multion,
  author    = {Wani, Saim and Patel, Shivansh and Jain, Unnat and Chang, Angel X. and Savva, Manolis},
  title     = {{MultiON}: Benchmarking Semantic Map Memory Using Multi-Object Navigation},
  booktitle = {Advances in Neural Information Processing Systems},
  volume    = {33},
  pages     = {9700--9712},
  year      = {2020},
}

@inproceedings{wang2026longterm,
  author    = {Wang, Sen and Liu, Bangwei and Gao, Zhenkun and Ma, Lizhuang and Wang, Xuhong and Xie, Yuan and Tan, Xin},
  title     = {Explore With Long-Term Memory: A Benchmark and Multimodal {LLM}-Based Reinforcement Learning Framework for Embodied Exploration},
  booktitle = {Proceedings of the {IEEE/CVF} Conference on Computer Vision and Pattern Recognition},
  pages     = {37098--37108},
  year      = {2026}
}

@inproceedings{werby2024hovsg,
  author    = {Werby, Abdelrhman and Huang, Chenguang and B{\"u}chner, Martin and Valada, Abhinav and Burgard, Wolfram},
  title     = {Hierarchical Open-Vocabulary 3D Scene Graphs for Language-Grounded Robot Navigation},
  booktitle = {Proceedings of Robotics: Science and Systems},
  address   = {Delft, Netherlands},
  year      = {2024},
  doi       = {10.15607/RSS.2024.XX.077},
}

@inproceedings{yenamandra2023homerobot,
  author    = {Yenamandra, Sriram and Ramachandran, Arun and Yadav, Karmesh and Wang, Austin S. and Khanna, Mukul and Gervet, Theophile and Yang, Tsung-Yen and Jain, Vidhi and Clegg, Alexander and Turner, John M. and Kira, Zsolt and Savva, Manolis and Chang, Angel X. and Chaplot, Devendra Singh and Batra, Dhruv and Mottaghi, Roozbeh and Bisk, Yonatan and Paxton, Chris},
  title     = {{HomeRobot}: Open-Vocabulary Mobile Manipulation},
  booktitle = {Proceedings of the 7th Conference on Robot Learning},
  series    = {Proceedings of Machine Learning Research},
  volume    = {229},
  pages     = {1975--2011},
  publisher = {PMLR},
  year      = {2023},
}

@misc{hiwonder2026jetrover,
  author       = {{Hiwonder}},
  title        = {{JetRover} {Jetson} Robot Car with AI Vision Robotic Arm},
  howpublished = {Online: \url{https://www.hiwonder.com/products/jetrover}},
  note         = {Accessed: Sep. 2, 2026},
}

@inproceedings{kohlbrecher2011hector,
  author    = {Kohlbrecher, Stefan and von Stryk, Oskar and Meyer, Johannes and Klingauf, Uwe},
  title     = {A Flexible and Scalable {SLAM} System with Full 3D Motion Estimation},
  booktitle = {2011 {IEEE} International Symposium on Safety, Security, and Rescue Robotics},
  pages     = {155--160},
  year      = {2011},
  doi       = {10.1109/SSRR.2011.6106777},
}

\end{document}